\documentclass[10pt,twocolumn,letterpaper]{article}

\usepackage{iccv}              
\usepackage{multirow}
\usepackage[accsupp]{axessibility}
\usepackage[normalem]{ulem}
\usepackage[table]{xcolor}
\usepackage{colortbl}
\definecolor{pastelviolet}{RGB}{230,220,250}
\definecolor{pastelyellow}{RGB}{255,250,205}
\usepackage[table]{xcolor}
\definecolor{pastelgreen}{RGB}{220,250,220}
\definecolor{pastelblue}{RGB}{220,235,250}
\definecolor{pastelpink}{RGB}{255,230,230}

\definecolor{iccvblue}{rgb}{0.21,0.49,0.74}
\usepackage[pagebackref,breaklinks,colorlinks,allcolors=iccvblue]{hyperref}

\input{preamble}

\def\paperID{8} 
\def\confName{ICCV}
\def\confYear{2025}

\title{Adapting Stereo Vision From Objects To 3D Lunar Surface Reconstruction with the StereoLunar Dataset }

\author{
Clémentine Grethen, Simone Gasparini, Géraldine Morin\\
IRIT, Toulouse INP, Université de Toulouse, France\\
{\tt\small \{clementine.grethen, simone.gasparini, geraldine.morin\}@irit.fr}
\and
Jérémy Lebreton, Lucas Marti\\
Airbus Defence and Space\\
{\tt\small \{jeremy.lebreton, lucas.marti\}@airbus.com}
\and
Manuel Sanchez Gestido\\
ESA (European Space Agency) \\
{\tt\small Manuel.Sanchez.Gestido@esa.int}
}

\begin{document}
\twocolumn[{
    \maketitle
    
   \begin{center}
    \includegraphics[width=.8\textwidth]{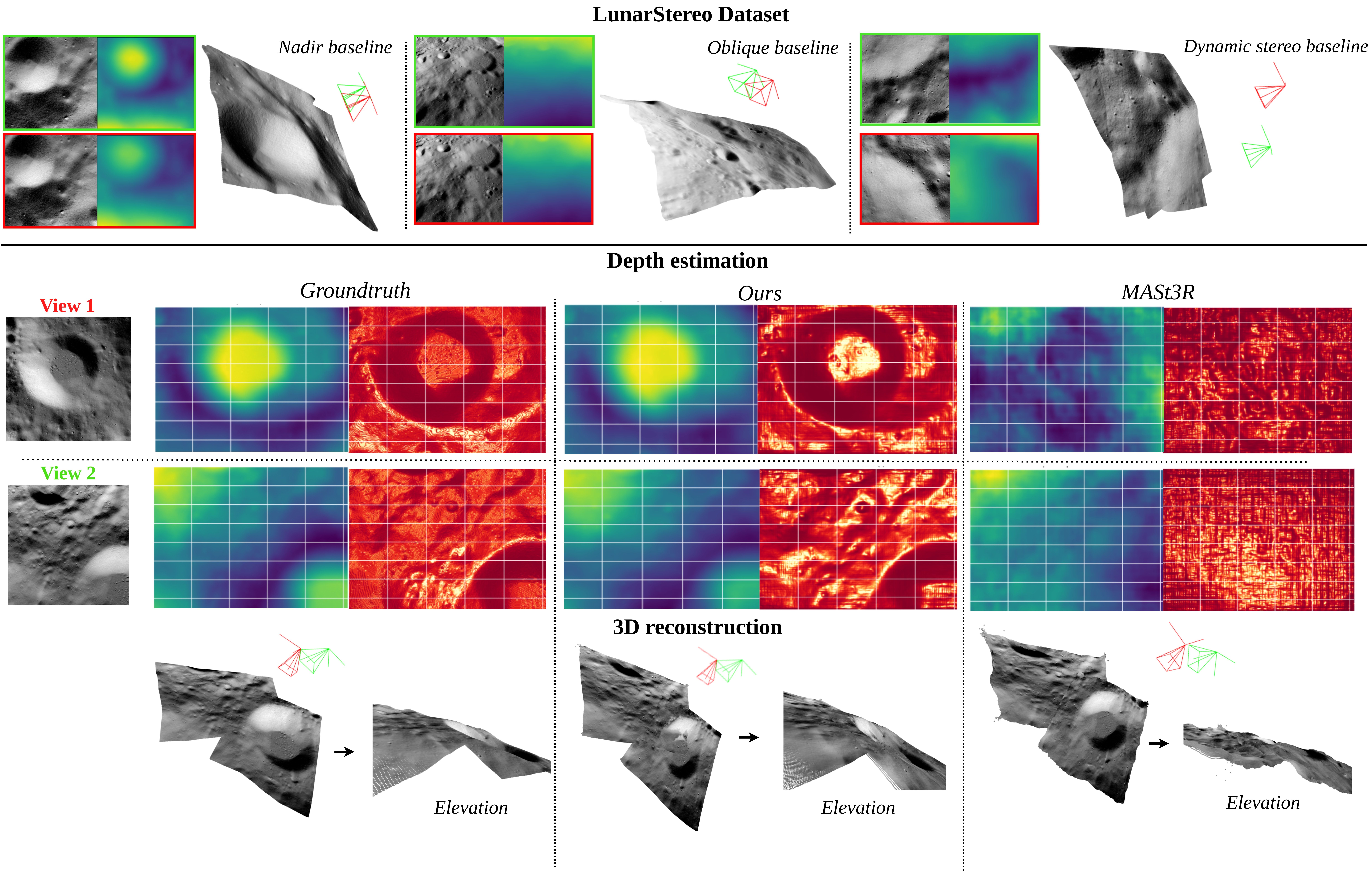}
    
\captionof{figure}{
\textit{\textbf{Top row}:} examples of lunar stereo image pairs of our dataset taken from three different trajectories, along with the corresponding ground-truth depth maps (viridis colormap) and 3D scene renderings.  
\textit{\textbf{Middle row}:} for each view, it shows the ground truth for the depth map (viridis colormap) and the slope map (heat colormap), followed by the corresponding predictions of our method and the MASt3R baseline.
\textit{\textbf{Bottom row}:} the ground truth 3D scene (left) with the final 3D reconstructions for our method (center) and MASt3R (right), with View 1 in \textcolor{red}{red} and View 2 in \textcolor{green}{green}. }

\label{fig:teasing}
\end{center}
    \vspace{1.5em}

    \vspace{1em} 
}]

\begin{abstract}
Accurate 3D reconstruction of lunar surfaces is essential for space exploration. 
However, existing  stereo vision reconstruction methods struggle in this context due to the Moon’s lack of texture, difficult lighting variations, and atypical orbital trajectories. 
State-of-the-art deep learning models, trained on human-scale datasets, have rarely been tested on planetary imagery and cannot be transferred directly to lunar conditions. 
To address this issue, we introduce \emph{LunarStereo}, the first open dataset of photorealistic stereo image pairs of the Moon, simulated using ray tracing based on high-resolution topography and reflectance models. 
It covers diverse altitudes, lighting conditions, and viewing angles around the lunar South Pole, offering physically grounded supervision for 3D reconstruction tasks. 
Based on this dataset, we adapt the MASt3R model to the lunar domain through fine-tuning on LunarStereo.
We validate our approach through extensive qualitative and quantitative experiments on both synthetic and real lunar data, evaluating 3D surface reconstruction and relative pose estimation. 
 
Extensive experiments on synthetic and real lunar data validate the approach, demonstrating significant improvements over zero-shot baselines and paving the way for robust cross-scale generalization in extraterrestrial environments.
\end{abstract}

\sisetup{group-separator = {,}}
\section{Introduction \& Background}
\label{sec:intro}
\noindent
Accurate 3D reconstruction of planetary surfaces is critically needed in space exploration missions, particularly for descent and landing, to perform Hazard Detection and Avoidance (HDA), trajectory planning, and autonomous navigation. 
For the Moon, the lack of an absolute Global Navigation Satellite System (GNSS), sparse visual features, and challenging lighting conditions make perception tasks very challenging. 
Thus, reliable and dense surface reconstruction is essential for landing a spacecraft safely on the Moon. 
In addition, reconstructing 3D models from images acquired during past or ongoing missions is key to building accurate topographic maps, refining prior knowledge of the terrain, and mission planning. 

Reconstruction during a landing sequence usually relies on classical computer vision pipelines, combining Structure-from-Motion (SfM) for sparse pose estimation with Multi-View Stereo (MVS) for dense depth reconstruction\cite{LeMoulic2024,Getchius2024}.
However, these methods are not well-suited to the constraints of spaceborne imagery. 
Their effectiveness is often limited in environments characterized by repetitive textures, low albedo variation, minimal stereo baseline, and strong illumination contrasts — all common in lunar imagery. 

Indeed, lunar imagery presents significant challenges. The surface is visually sparse, highly repetitive, and exhibits fractal-like patterns, with minimal color variation due to the absence of an atmosphere, resulting in low contrast at high altitude and significant difficulties for visual interpretation and image-based algorithms \cite{Getchius2024,Posada2024,kumar2024moonmetasynclunarimageregistration}.
Furthermore, the descent trajectories are often near-nadir, thus introducing a degenerate configuration for monocular SfM due to the lack of lateral parallax~\cite{Sturm1997Critical}.
Shadows and lighting gradients further affect feature detection and 3D estimation. 
As recent autonomous landing failures have shown~\cite{Foust_2025}, the ability to perceive and understand the terrain in such conditions remains a critical open problem.
In recent years, deep learning-based approaches have emerged as a means of overcoming some of the limitations of classical 3D reconstruction pipelines. 
These include learned feature matchers such as SuperGlue~\cite{sarlin2020superglue}, end-to-end stereo networks such as MVSNet~\cite{yao2018mvsnet}. 
More recently, unified architectures such as MASt3R~\cite{leroy2024grounding}, DUSt3R~\cite{wang2023dust3r}, and VGGT~\cite{wang2025vggt} has been introduced for the 3D reconstruction from general images: trained on large-scale terrestrial datasets like MegaDepth~\cite{MDLi18} and Co3Dv2~\cite{reizenstein21co3d}, these models achieve state-of-the-art performance on human-scale imagery — including urban scenes, natural landscapes, indoor environments, and common objects. 
However, their effectiveness remains largely limited to the types of scenes represented in their training data. When applied to out-of-domain settings such as lunar imagery, their performance degrades significantly \cite{vuong2025aerialmegadepthlearningaerialgroundreconstruction}. 
For example, applying MASt3R to raw lunar images often results in unreliable geometry, with flat reconstructions, inconsistent relief, or noisy outputs, as illustrated in \cref{fig:teasing}. 
This highlights a clear domain gap: deep models trained on Earth-like content do not generalize well to the low-texture, high-altitude, and structured viewpoints found in space applications — unless carefully adapted.

To address this limitation, we introduce the first publicly available physically realistic lunar stereo dataset, \textit{LunarStereo}, tailored for deep learning-based 3D reconstruction. 
Our dataset includes simulated stereo pairs generated through ray tracing over high-resolution Digital Elevation Models (DEMs), using accurate reflectance models (BRDFs), realistic Sun illumination, and varied camera trajectories. 
This enables us to provide dense ground-truth depth maps and accurate camera poses under a wide range of imaging conditions, including nadir and oblique views, low altitudes, and challenging illumination, closely mimicking real descent scenarios (\cf first row of \cref{fig:teasing} for different samples of the dataset). 

We then used this dataset to fine-tune the MASt3R model for lunar stereo vision.
We chose MASt3R due to its ability to output 3D correspondences, its superior performance compared to DUST3R, and its significantly lighter architecture compared to VGGT.
Our fine-tuned version shows significant improvements in reliably reconstructing the surface and, in particular, the geometry of the reliefs, as can be observed in \cref{fig:teasing}. 
These findings demonstrate not only the potential of MASt3R in space imaging but also, more broadly, the adaptability of modern 3D vision networks to low-texture and out-of-distribution domains through targeted fine-tuning.

\noindent\textbf{The contributions of this work are twofold:}
(i) We release the first publicly available, high-fidelity, lunar stereo dataset with full geometric supervision, simulated from DEMs under physically-based rendering. \footnote{Data and code: \url{https://clementinegrethen.github.io/publications/3D-Vast-ICCV2025.html}}  
While high-frequency texture details are not modeled, this enables wide coverage and flexible illumination control beyond real conditions at the South Pole. 
(ii) We fine-tune the MASt3R model on this dataset, demonstrating its successful adaptation to the lunar domain with dramatically improved 3D geometry estimation across all scenarios, achieving an average reduction of over \SI{70}{\percent} in slope estimation error and significantly enhancing overall relative accuracy by roughly \SI{50}{\percent}.

\noindent
The rest of the paper is organized as follows: \cref{sec:related_work} reviews prior work on lunar datasets and 3D reconstruction methods. \cref{sec:dataset_generation} details our dataset creation pipeline. \cref{sec:pose_geometry} describes the reconstruction and fine-tuning methodology. \cref{sec:experiments} presents our evaluation and results. Finally, \cref{sec:conclusion} discusses limitations and future directions.
\section{Related work}
\label{sec:related_work}
\subsection{Lunar images datasets}
To evaluate and adapt deep learning–based 3D reconstruction methods to the lunar environment, we require datasets that combine imagery with accurate geometric supervision, stereo pairs or multi-view images with consistent camera metadata, dense geometry, and sufficient diversity in lighting, viewpoint, and terrain. 
In this subsection, we review the main publicly available lunar resources and assess their suitability for such tasks. 
We consider four categories: (1) Digital Elevation Models (DEMs), which serve as geometric references and enable synthetic rendering; (2) real lunar image datasets; (3) synthetic datasets generated from simulations; and (4) laboratory-controlled datasets with ground-truth geometry.

\myparagraph{Lunar Elevation Models}  
Several DEMs of the Moon are publicly available, with varying resolution, coverage, and acquisition methods. \cref{tab:lunar_dems} summarizes the most widely used one. 
Global models from LOLA (aboard NASA’s Lunar Reconnaissance Orbiter, LRO)~\cite{BARKER2016346}, Kaguya-LOLA (JAXA)~\cite{Araki2009}, and Chang'E-2 (CNSA)~\cite{1671-8860(2018)04-0485-11} provide consistent terrain baselines at different scales. 
In addition, local high-resolution DEMs at \SIrange{2}{5}{\meter} resolution have been produced from stereo pairs acquired by LRO’s Narrow Angle Camera (NAC)\cite{Robinson2010}. 
These tiles offer the most detailed public lunar topography, but they remain limited in number and spatial extent, mostly targeting landing sites and selected scientific regions.

\begin{table}[h]
\centering
\caption{Main publicly available lunar DEMs.}
\label{tab:lunar_dems}
\setlength{\tabcolsep}{3pt}  
\renewcommand{\arraystretch}{1.1}  
\footnotesize  

\begin{tabular}{llll}
\toprule
\textbf{Mission} & \textbf{Coverage} & \textbf{Res.} & \textbf{Method} \\
\midrule
LOLA SLDEM2015~\cite{Barker2016}       & Global       & \SI{118}{\meter}   & Laser altimetry \\
LOLA (Pole)~\cite{Barker2021}       & South pole   & \SI{5}{\meter}     & Dense tracks \\
Kaguya (JAXA)~\cite{Araki2009}     & Global       & \SI{59}{\meter}    & Stereo imagery \\
Chang'E-2 (CNSA)~\cite{1671-8860(2018)04-0485-11}  & Global       & \SI{20}{\meter}    & Pushbroom stereo \\
LRO NAC-derived~\cite{Robinson2011LROCRDR}  & Sparse sites & \SIrange[range-phrase={--}]{2}{5}{\meter}  & From stereo pairs \\
\bottomrule
\end{tabular}
\end{table}

\myparagraph{Real Lunar Image Datasets}
 Different real image datasets can be used to validate lunar vision tasks, but they vary in resolution and often lack ground truth camera pose and intrinsic parameters, making geometric supervision inaccurate. 
 The Chang'E Lunar Landscape~\cite{h8v1-rm55-24} includes over \num{7500} images from the Chang'E-3 and Chang'E-4 landers' descent sequences, but it provides no accurate 3D ground truth. 
 In contrast, datasets from orbital imagery provide different challenges. 
 Chandrayaan-2 offers images, including stereo triplets, at higher resolutions of $\SI{\approx30}{\centi\meter\per\px}$~\cite{issdcChandrayaanII}, but without per-image calibration, camera poses, or associated terrain models. 
 NASA's LRO NAC delivers panchromatic orbital strips down to a resolution of $\SI{\approx 0.5}{\meter\per\px}$, but their narrow field of view and sparse coverage prevent their use in standard MVS or SfM pipelines. 
 More broadly, most available datasets are constrained to nadir orbital imaging, with limited variation in viewpoint, motion, or camera baseline. 
 This lack of geometric diversity, even when terrain models are available, makes it difficult to train or evaluate learning-based 3D methods. 
 To date, no publicly available dataset provides true lunar stereo imagery with accurate poses, intrinsics, and dense ground-truth geometry.

\myparagraph{Synthetic simulation datasets.}
To address the scarcity of real lunar imagery, synthetic datasets are often generated using common commercial rendering engines such as Unreal Engine, Terragen, or tools specifically developed for rendering space images, such as PANGU~\cite{Parkes2004} and SurRender~\cite{lebreton2022high}.
These offer diverse viewpoints and rich metadata, including pixel-level ground truth. 
The Artificial Lunar Landscape~\cite{romain_pessia_prof__genya_ishigami_quentin_jodelet_2019} contains thousands of labeled images for semantic tasks but no geometrical ground truth. LuSNAR~\cite{liu2024lusnaralunarsegmentationnavigation} provides photorealistic stereo pairs, depth maps, and multi-sensor data generated with Unreal Engine, making it attractive for SLAM and stereo-based learning. 
However, LuSNAR lacks a realistic physical model of the Moon, as it does not account for real lunar terrain, reflectance properties, or solar interactions. Moreover, it only includes ground-level trajectories, limiting viewpoint diversity. 

More broadly, synthetic datasets fail to reproduce the Moon’s physical realism, especially in terms of BRDF, terrain variability, and lighting conditions, raising concerns about their generalization to real lunar imagery. 

\myparagraph{Laboratory-Controlled Datasets}
Laboratory-controlled datasets are obtained using mock-ups and camera trajectories controlled by robotic arms, thus offering a precise ground truth.  
DLR’s TRON dataset~\cite{lebreton2024trainingdatasetsgenerationmachine} includes \num{7238} images captured over a \qtyproduct{4 x 2}{\meter} mock-up, but the 3D ground truth is not publicly available at the moment.
NASA’s POLAR dataset~\cite{wong2017polar} provides 
\num{2500} HDR stereo pairs over a small regolith simulant scene in the Moon's South pole, centered on a single crater and a few scattered rocks.
However, both datasets have limited spatial extent and low landscape diversity. 
TRON uses artificial surfaces and indirect lighting, differing from lunar reflectance and solar conditions. 
POLAR targets only polar regions of the Moon, where the sunlight arrives at very shallow/tangential angle, thus resulting in low light conditions and very dark imagery.
As a result, these datasets may lead to models overfitting to specific textures or terrain structures, far from the variability and appearance of real lunar landscapes.

To the best of our knowledge, no existing publicly available dataset provides stereo lunar imagery that jointly integrates physical coherence, such as realistic lunar appearance and extensive variation in displacement, illumination, altitude, and landscape diversity.

\subsection{3D Reconstruction in Lunar Context}
3D reconstruction techniques have been increasingly applied to planetary exploration missions, motivated by the need for autonomous navigation, geological analysis, and large-scale terrain modeling from orbital or descent imagery. 
However, reconstructing accurate 3D geometry from lunar images remains challenging due to low-texture surfaces or illumination variations.

Several photometric methods, especially Shape-from-Shading~\cite{Alexandrov2018} and photometric stereo~\cite{Nefian2013,Peng2021}, have been proposed to recover surface geometry from illumination-driven image variations. 
While effective in some scenarios, these methods typically assume fixed or known illumination and are highly sensitive to the Moon’s reflectance properties. 
As such, they do not generalize well to wide-baseline multiview reconstruction from unconstrained orbital or descent imagery, which is the focus of our study.

Structure-from-Motion (SfM) pipelines reconstruct sparse 3D geometry by identifying correspondences across multiple views and estimating camera motion. 
In the context of the Moon, these methods face significant challenges. 
Image registration is particularly difficult due to the low-texture and repetitive nature of the terrain, as well as strong illumination gradients. 
Even recent efforts using custom detectors and interpolation~\cite{kumar2024moonmetasynclunarimageregistration} show limited gains over classical approaches. 
To address this issue, it has been shown that learning-based matchers such as DISK combined with LightGlue~\cite{Posada2024} can provide more robust and dense correspondences under harsh lunar conditions.
In addition, descent trajectories are often constrained by the mission, and they are frequently nadir-oriented (\eg Chang'E3 landing), thus providing  an insufficient baseline for reliable triangulation.
SfM pipelines are also highly sensitive to illumination variations, which further degrades matching quality across views.
These limitations were evident in the Nova-C mission~\cite{freeman2025soft,Getchius2024}, whose descent trajectory was explicitly designed to support stereo-based hazard detection and terrain reconstruction. 
This highlights the need for dedicated image acquisition configurations to make classical SfM viable. 

In contrast, our study tackles a broader range of stereo displacements, including unconstrained descent sequences, specifically aiming to overcome the limitations of classical methods. 
We achieve this by leveraging deep learning–based correspondence methods that are more robust to viewpoint and illumination variations. 
Recent end-to-end models such as DUSt3R~\cite{wang2023dust3r}, MASt3R~\cite{leroy2024grounding}, and VGGT~\cite{wang2025vggt} have shown strong performance on terrestrial scenes by jointly learning correspondences, poses, and dense geometry.
Aerial MegaDepth~\cite{vuong2025aerialmegadepthlearningaerialgroundreconstruction} demonstrated the adaptation of such models to aerial-to-ground scenarios using pseudo-synthetic city-scale data. 
However, this setting differs from ours: lunar imagery is characterized by sparse texture, low-frequency terrain, and extreme lighting, elements not present in urban datasets. 
To the best of our knowledge, these deep learning networks have not yet been thoroughly tested on the challenging context of lunar imagery reconstruction.

\section{Our Proposed Lunar Stereo Dataset}
\label{sec:dataset_generation}

\begin{table*}[ht]
\centering
{\scriptsize
\begin{tabular}{lcccccccccc}
\toprule
\textbf{Altitude (km)} & 3.5 & 6.2 & 9.5 & 12.8 & 16.1 & 19.4 & 22.7 & 26.0 & 29.2 & 30.5 \\
\textbf{Effective GSD (\si{\meter\per\px})} & 5.7 & 10.0 & 15.4 & 20.7 & 26.0 & 31.4 & 36.7 & 42.1 & 47.2 & 49.3 \\
\bottomrule
\end{tabular}
}
\caption{Altitude vs. effective ground sampling distance (GSD).}
\label{tab:gsd_vs_alt}
\end{table*}

\begin{figure*}[t]
    \centering
    \includegraphics[width=.7\linewidth]{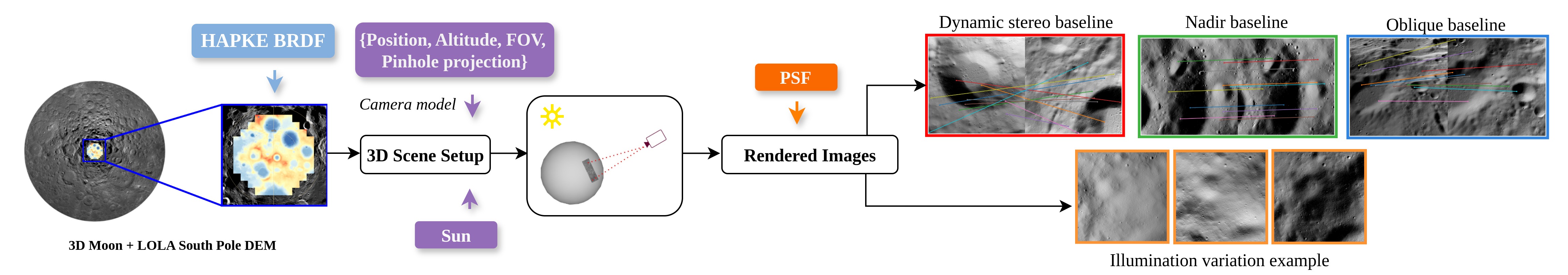}
    \caption{Overview of the dataset generation pipeline. 
    The output is images captured from the three trajectory types, and offers ground truth pixel-level correspondences. 
    For each pair, three variants with different illumination configurations have been generated.}

    \label{fig:rendu}
\end{figure*}

In this section, we present the first publicly available dataset of stereo lunar images designed for learning-based 3D reconstruction. 
Our dataset covers diverse terrains and lighting, provides per-pixel depth, full camera metadata, and physically based rendering using Moon's bidirectional reflectance distribution function (BRDF).
As illustrated in \cref{fig:rendu}, our rendering pipeline generates physically accurate stereo pairs by combining: (1) a high-resolution lunar DEM, (2) a BRDF reflectance model, (3) different realistic sun  illuminations, and (4) a parametric camera model. 
These are integrated in the \textit{SurRender} ray tracing engine~\cite{lebreton2022high} to produce geometrically accurate image pairs with full 3D supervision.

\myparagraph{Lunar Terrain and Illumination Modeling.}  
As terrain input, we use the LOLA South Pole DEM at \SI{5}{\meter\per\px}. 
This dataset, derived from the laser altimeter on board NASA’s LRO mission, provides high vertical accuracy and detailed coverage of polar craters. 
It enables us to simulate varied landscapes, from illuminated ridges to deep, permanently shadowed basins, with elevation values ranging from \SI{-4350}{\meter} to \SI{+1850}{\meter} (relative to lunar mean radius)\footnote{Coarser global DEMs (\eg, Chang'E-2 at \SI{20}{\meter\per\px}) are insufficient to capture such variations.}.
To simulate surface reflectance, we adopt the Hapke BRDF~\cite{hapke1993theory}, a physically grounded model tailored for airless, regolith-covered bodies.
This BRDF accounts for the Moon's unique reflectance phenomena, such as the opposition effect and anisotropic scattering. 
While global albedo maps are too coarse for our 512 × 512 image patches, we assume constant albedo to avoid high-frequency artefacts.
Lighting is simulated by varying solar azimuth and elevation to reflect the diverse and low-angle conditions typical at the South Pole. 
This setup produces strong cast shadows and photometric variation across viewpoints as depicted in  \cref{fig:rendu}).

\myparagraph{Stereo Rendering and Camera Simulation.}  
The cameras are modeled as a pinhole projection model with known intrinsics and 6-DOF extrinsics. 
Each stereo pair is rendered at a fixed field of view (\ang{45}) and resolution (\qtyproduct{512 x 512}{\px}), with optical blur physically simulated by sampling rays according to a Gaussian Point-Spread Function (PSF), as part of the rendering process. 
We simulate three types of camera motion, inspired by the typical phases observed during lunar descent. 
To guide the design of realistic configurations, we qualitatively analyzed the trajectory data from the Chang'E-3 mission and other descent studies such as Nova-C~\cite{YU2014,Wong2006,Getchius2024}. 
These motion patterns are designed to capture diverse observation geometries while ensuring a sufficient parallax for stereo matching. 
All parameters were empirically tuned through iterative experimentation to balance realism, geometric diversity, and reconstruction difficulty:
\begin{itemize}
    \item \textbf{Nadir:} The cameras look vertically down, simulating controlled vertical descent. Baselines range from \SIrange{4}{10}{\percent} of the altitude, with both cameras at the same height.
    
    \item \textbf{Oblique:} Cameras are tilted, with viewing angles between \ang{20} and \ang{35}, reflecting lateral motion or target-centered reorientations. 
    We vary the altitudes of the stereo pair in different ways: (1) the cameras are placed at the same altitude, (2) one is slightly or significantly higher than the other.
    
    \item \textbf{Dynamic:} A more challenging case with additional variation: camera altitudes vary by up to \SI{\pm 30}{\percent}, roll by \ang{\pm 10}, and viewpoints are oblique or near-nadir. 
    The baselines are randomized from \SIrange{5}{18}{\percent} of the altitude to increase diversity.
\end{itemize}

\noindent Stereo pairs are generated across $10$ altitude bands, from  \SI{3.5}{\kilo\metre} to \SI{30.5}{\kilo\metre}, around which stereo pairs are generated with small variations. 
These altitude levels affect the ground sampling distance (GSD), as summarized in \cref{tab:gsd_vs_alt}.

Each trajectory is rendered under three distinct lighting conditions, chosen to produce different shadow patterns and levels of darkness. 
We vary the Sun's azimuth and incidence angles to simulate side (\ang{150}, \ang{160}), overhead (\ang{250}, \ang{20}), and back lighting (\ang{360}, \ang{165}). 
These setups allow the same scene to reveal different geometric cues under changing illumination, as illustrated in \cref{fig:rendu}.

\myparagraph{Ground Truth Parameters.}
Each stereo pair is provided with:
\begin{itemize}
    \item \textbf{Intrinsic parameters:} focal length, principal point, sensor size;
    \item \textbf{Extrinsics:} camera-to-world poses in Moon-fixed Frame;
    \item \textbf{Dense depth maps:} per-pixel depth along camera rays;
    \item \textbf{Stereo baseline:} 3D inter-camera displacement and rotation;
    \item \textbf{Georeferenced trajectory metadata:} including absolute altitude and GSD.
\end{itemize}

Finally, camera positions are uniformly sampled across the lunar south polar cap (from \ang{-90} to \ang{-87}) and the full longitude range to ensure spatial coverage and diversity. 
The resulting dataset comprises over \num{50000} stereo pairs, well distributed across a range of altitudes, illumination conditions, terrains, and camera trajectories. 
This enables reproducible and high-fidelity benchmarks for stereo vision in realistic lunar settings, supporting future research in 3D reconstruction, terrain analysis, and autonomous or crewed lunar exploration. The dataset will be publicly released to foster further development and evaluation in the community.

\section{Learning Moon 3D Reconstruction }
\label{sec:pose_geometry}
\begin{table*}[!htb]
    \centering
    \resizebox{\textwidth}{!}{%
    \begin{tabular}{l|
>{\columncolor{pastelgreen}}c >{\columncolor{pastelgreen}}c >{\columncolor{pastelgreen}}c >{\columncolor{pastelgreen}}c |
>{\columncolor{pastelblue}}c >{\columncolor{pastelblue}}c >{\columncolor{pastelblue}}c >{\columncolor{pastelblue}}c |
>{\columncolor{pastelpink}}c >{\columncolor{pastelpink}}c >{\columncolor{pastelpink}}c >{\columncolor{pastelpink}}c |
>{\columncolor{pastelgreen}}c >{\columncolor{pastelgreen}}c >{\columncolor{pastelgreen}}c >{\columncolor{pastelgreen}}c |
>{\columncolor{pastelblue}}c >{\columncolor{pastelblue}}c >{\columncolor{pastelblue}}c >{\columncolor{pastelblue}}c |
>{\columncolor{pastelpink}}c >{\columncolor{pastelpink}}c >{\columncolor{pastelpink}}c >{\columncolor{pastelpink}}c}
        \toprule
        \multirow{2}{*}{\textbf{Method}} & \multicolumn{12}{c|}{\textbf{RRA (\% below threshold)}} & \multicolumn{12}{c}{\textbf{RTA (\% below threshold)}} \\
        \cmidrule(lr){2-13} \cmidrule(lr){14-25}
        & \multicolumn{4}{c|}{\textbf{Nadir}} & \multicolumn{4}{c|}{\textbf{Oblique}} & \multicolumn{4}{c|}{\textbf{Dynamic}}
        & \multicolumn{4}{c|}{\textbf{Nadir}} & \multicolumn{4}{c|}{\textbf{Oblique}} & \multicolumn{4}{c}{\textbf{Dynamic}} \\
        \cmidrule(lr){2-5} \cmidrule(lr){6-9} \cmidrule(lr){10-13}
        \cmidrule(lr){14-17} \cmidrule(lr){18-21} \cmidrule(lr){22-25}
        & $2^\circ$ & $5^\circ$ & $15^\circ$ & $30^\circ$
        & $2^\circ$ & $5^\circ$ & $15^\circ$ & $30^\circ$
        & $2^\circ$ & $5^\circ$ & $15^\circ$ & $30^\circ$
        & $2^\circ$ & $5^\circ$ & $15^\circ$ & $30^\circ$
        & $2^\circ$ & $5^\circ$ & $15^\circ$ & $30^\circ$
        & $2^\circ$ & $5^\circ$ & $15^\circ$ & $30^\circ$ \\
        \midrule
        ORB     & 84.7 & 99.0 & \textbf{100.0} & \textbf{100.0} & 92.0 & 96.3 & 97.3 & 98.3 & 70.0 & 88.0 & 96.8 & 97.8 & 19.0 & 43.3 & 72.3 & 78.0 & 44.7 & 79.7 & 93.3 & 96.0 & 27.8 & 59.2 & 84.8 & 90.5 \\
        MASt3R  & 98.7 & 99.7 & \textbf{100.0} & \textbf{100.0} & \textbf{100.0} & \textbf{100.0} & \textbf{100.0} & \textbf{100.0} & 90.1 & 92.4 & 94.4 & 95.4 & 56.6 & 82.5 & \textbf{94.3} & 97.0 & 95.9 & \textbf{99.3} & \textbf{100.0} & \textbf{100.0} & 76.5 & 89.8 & 93.0 & 95.1 \\
        Ours    & \textbf{98.3} & \textbf{100.0} & \textbf{100.0} & \textbf{100.0} & \textbf{100.0} & \textbf{100.0} & \textbf{100.0} & \textbf{100.0} & \textbf{98.0} & \textbf{98.5} & \textbf{99.2} & \textbf{99.3} & \textbf{57.2} & \textbf{84.5} & 93.3 & \textbf{97.3} & \textbf{96.9} & 99.0 & 99.7 & \textbf{100.0} & \textbf{91.7} & \textbf{96.6} & \textbf{98.3} & \textbf{99.5} \\
        \bottomrule
    \end{tabular}
    }
    \caption{Proportion of image pairs with RRA and RTA below specified angular thresholds (2, 5, 15, 30), comparing methods across datasets. Nadir columns are highlighted in green, oblique in blue, and dynamic in pink. Best scores are \textbf{bold}.}
    \label{tab:rra_rta_by_metric_dataset_ultracompact}
\end{table*}

\begin{table*}[ht]
\centering
\caption{
Compact metrics for MASt3R and our method across different datasets.
Arrows indicate whether higher (↑) or lower (↓) values are better.
Best scores are \textbf{bold}.
Columns highlighted in \colorbox{pastelviolet}{\textbf{violet}} (bottom table) represent terrain-related metrics (slope correlation, slope MAE (in meters), SSIM, depth profile correlation),
while columns highlighted in \colorbox{pastelyellow}{\textbf{yellow}} (top table)  correspond to standard 3D reconstruction metrics (accuracy, completeness, overall Chamfer distance).
}

\vspace{0.5em}

\scalebox{0.75}{
\setlength{\tabcolsep}{2pt}
\begin{tabular}{l|
>{\columncolor{pastelyellow}}c >{\columncolor{pastelyellow}}c >{\columncolor{pastelyellow}}c |
>{\columncolor{pastelyellow}}c >{\columncolor{pastelyellow}}c >{\columncolor{pastelyellow}}c |
>{\columncolor{pastelyellow}}c >{\columncolor{pastelyellow}}c >{\columncolor{pastelyellow}}c}
\toprule
\textbf{Method} & \multicolumn{3}{c|}{\textbf{Nadir}} & \multicolumn{3}{c|}{\textbf{Oblique}} & \multicolumn{3}{c}{\textbf{Dynamic}} \\
\cmidrule(lr){2-4} \cmidrule(lr){5-7} \cmidrule(lr){8-10}
& ACC.(m/rel) $\downarrow$ & Compl.(m/rel) $\downarrow$ & Chamfer(m/rel) $\downarrow$
& ACC. (m/rel) $\downarrow$ & Compl.(m/rel) $\downarrow$ & Chamfer(m/rel) $\downarrow$
& ACC. (m/rel) $\downarrow$ & Compl.(m/rel) $\downarrow$ & Chamfer(m/rel) $\downarrow$ \\
\midrule
MASt3R & 236 / 1.10\% & 235 / 1.06\% & 236 / 1.08\%
       & 385 / 1.12\% & 259 / 0.75\% & 322 / 0.93\%
       & 289 / 1.10\% & 270 / 1.18\% & 279 / 1.14\% \\
Ours   & \textbf{103 / 0.47\%} & \textbf{97 / 0.45\%} & \textbf{100 / 0.48\%}
       & \textbf{141 / 0.47\%} & \textbf{147 / 0.49\%} & \textbf{144 / 0.48\%}
       & \textbf{109 / 0.40\%} & \textbf{114 / 0.41\%} & \textbf{111 / 0.41\%} \\
\bottomrule
\end{tabular}
}

\vspace{1em}

\scalebox{0.75}{
\setlength{\tabcolsep}{2.3pt}
\begin{tabular}{l|
>{\columncolor{pastelviolet}}c >{\columncolor{pastelviolet}}c >{\columncolor{pastelviolet}}c >{\columncolor{pastelviolet}}c |
>{\columncolor{pastelviolet}}c >{\columncolor{pastelviolet}}c >{\columncolor{pastelviolet}}c >{\columncolor{pastelviolet}}c |
>{\columncolor{pastelviolet}}c >{\columncolor{pastelviolet}}c >{\columncolor{pastelviolet}}c >{\columncolor{pastelviolet}}c}
\toprule
\textbf{Method} & \multicolumn{4}{c|}{\textbf{Nadir}} & \multicolumn{4}{c|}{\textbf{Oblique}} & \multicolumn{4}{c}{\textbf{Dynamic}} \\
\cmidrule(lr){2-5} \cmidrule(lr){6-9} \cmidrule(lr){10-13}
& Slope corr. $\uparrow$ & Prof MAE $\downarrow$ & SSIM $\uparrow$ & Prof corr. $\uparrow$
& Slope corr. $\uparrow$ & Prof MAE $\downarrow$ & SSIM $\uparrow$ & Prof corr. $\uparrow$
& Slope corr. $\uparrow$ & Prof MAE $\downarrow$ & SSIM $\uparrow$ & Prof corr. $\uparrow$ \\
\midrule
MASt3R & 0.21 & 218.53 & 0.31 & 0.76
       & 0.39 & 197.77 & 0.51 & 0.88
       & 0.09 & 229.44 & 0.24 & 0.63 \\
Ours   & \textbf{0.80} & \textbf{46.77} & \textbf{0.78} & \textbf{0.97}
       & \textbf{0.82} & \textbf{57.71} & \textbf{0.83} & \textbf{0.99}
       & \textbf{0.76} & \textbf{57.83} & \textbf{0.76} & \textbf{0.95} \\
\bottomrule
\end{tabular}
}
\label{tab:metrics_split}
\end{table*}

\begin{figure*}[!htb]
    \centering
    \includegraphics[width=0.95\linewidth]{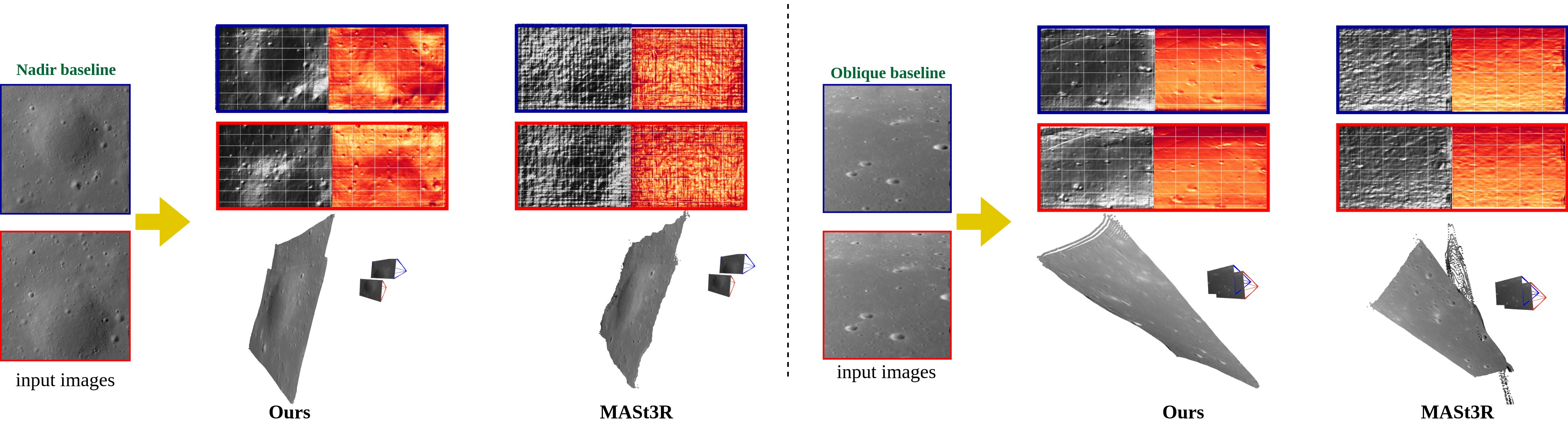}
    \caption{Qualitative study using real descent imagery. Comparison between \textit{MASt3R} (pretrained) and \textit{Ours} on a Nadir baseline (left column) and an Oblique baseline (right column). {Additional real-world examples are provided in the \textbf{supplementary material}.}}

    \label{fig:qualitative-real}
\end{figure*}
We explore the impact and benefits of our LunarStereo dataset on supervised learning for multi-view 3D reconstruction.

\myparagraph{MASt3R Architecture}
We consider the problem of estimating extrinsic camera parameters and the 3D structure of the lunar surface from two stereo images. 
To this end, we fine-tune the MASt3R model~\cite{leroy2024grounding}, which jointly performs 3D reconstruction and feature matching. 
A central component of MASt3R's 3D reconstruction process is the pointmap representation, denoted $\mathcal{P}$. 
For a given input image $I^a$, the model predicts a dense 2D-to-3D mapping to a 3D point cloud $X^{a,b} \in \mathbb{R}^{H \times W \times 3}$ expressed in the coordinate system of a reference camera $C^b$
This pointmap encodes the 3D scene geometry for every pixel. 

In addition to 3D regression, MASt3R includes a descriptor head that predicts dense local feature maps, $D^1, D^2 \in \mathbb{R}^{H \times W \times d}$, optimized for accurate pixel-level matching. 
From these features, reliable correspondences can be established between the two input images.

These correspondences enable the estimation of relative camera poses through two approaches: (1) by computing the essential matrix from 2D–2D matches or (2) by applying Perspective-n-Point (PnP)\cite{Grunert1841} using 2D–3D matches derived from the predicted pointmap $\mathcal{P}^2$ of the second image. 
The latter recovers the full relative transformation $\mathbf{T} = [\mathbf{R} \,|\, \mathbf{t}]$ between the two views. This integration of feature matching and geometric estimation within a single framework yields strong performance for pose prediction and 3D reconstruction. 
Our goal is to benefit from the MASt3R architecture and performance for lunar reconstruction.

\myparagraph{Fine-tuning details}
We initialize the model with the publicly available MASt3R checkpoints~\cite{wang2023dust3r}, which were pre-trained on a large mixture of $14$ diverse datasets featuring millions of real-world and synthetic images, including indoor, outdoor, and object-centric scenes.
We then fine-tune the model on a selection of approximately \textbf{\num{31000}} image pairs from our training set, carefully chosen to ensure comprehensive coverage of position, lighting, altitude, and stereo displacement variations (uniform sampling in feature space), while keeping the encoder frozen.
We applied various data augmentation techniques to enhance robustness and mitigate overfitting, including color jitter, random cropping, grayscale tinting, and bilateral filtering to reduce texture and contrast. These pairs were split into an \SI{80}{\percent} training set and a \SI{20}{\percent} validation set.
The training is conducted for 25 epochs using the \textit{AdamW} optimizer with a learning rate of \num{3e-5} to mitigate overfitting, running on two NVIDIA Quadro RTX 8000 GPU, with a 2 batch size.

\myparagraph{Loss precision}
The Moon's fractal-like surface properties make it difficult to establish a consistent metric scale. For this reason, we do not require the network to learn a metric scale. 
Instead, we use the scale-invariant version of the regression loss, as defined in the original DUSt3R work:
\begin{equation}
    l_{regr}(\nu, i) = \left\| \frac{1}{z}X_{i}^{\nu,1} - \frac{1}{\hat{z}}\hat{X}_{i}^{\nu,1} \right\|,
\end{equation}
where normalizing factors $z$ and $\hat{z}$ are defined as the mean distance of all valid 3D points to the origin, making the reconstruction invariant to scale.

\section{Experiments}
\label{sec:experiments}

For the evaluation, we generated a new test set of \num{3000} stereo image pairs with altitudes interpolated between training values, using the same LOLA \SI{5}{\meter} DEM.
Finally, we perform a qualitative evaluation on real lunar images to illustrate applicability in real-world conditions.

\subsection{Pose estimation}
Following prior work~\cite{arnold2022mapfreevisualrelocalizationmetric,leroy2024grounding}, we evaluate camera pose estimation using Relative Rotation Accuracy (RRA) and Relative Translation Accuracy (RTA), which measure the angular error between predicted and ground-truth relative rotations and translation directions, respectively. 
We report RRA@$\tau$ and RTA@$\tau$, \ie, the percentage of camera pairs with error below a threshold $\tau$.

\cref{tab:rra_rta_by_metric_dataset_ultracompact} shows the results for the pose estimation obtained from the essential matrix estimation using 2D matches and known intrinsics. 
For comparison, we include the results of a classic baseline using ORB as features, as proposed in~\cite{Getchius2024}. 
As expected, the ORB-based approach performs poorly on RRA in the Dynamic configuration, where wide-baseline and viewpoint changes make feature matching challenging. 
For RTA, our method has, in general, better or comparable performances \wrt MASt3r, but it still shows some limitations in estimating the translation, especially in the Nadir sequence.

In general, our method achieves comparable, if not better, performance than MASt3R across most configurations. 
On the other hand, it consistently outperforms MASt3R on the Dynamic baseline, demonstrating stronger robustness to viewpoint and terrain variability.

\myparagraph{3D Reconstruction Evaluation Metrics}
To evaluate the accuracy of the 3D reconstruction, we follow the standard protocol used in MASt3R~\cite{leroy2024grounding}, reporting the Chamfer distance~\cite{Barrow1977Parametric}, accuracy, and completeness~\cite{Seitz2006Comparison}. 
Accuracy is defined as the average distance from each reconstructed 3D point to its closest ground-truth point; completeness measures the reverse; and the Chamfer distance is the average of both. 
These metrics quantify point-wise fidelity but do not guarantee structural consistency of the underlying surface morphology, particularly in large-scale, low-texture lunar terrain. 
To better assess whether the reconstruction preserves the true shape of the scene, we introduce a slope-based evaluation. 
This is a classical terrain feature used in landing safety assessments~\cite{Steffes2023}, as it directly relates to surface stability. 
We compute per-pixel slopes directly from the predicted and ground-truth 3D point maps, using spatial finite differences on the elevation channel. 
Local surface gradients are then combined to derive slope angles, from which we compute the Pearson correlation between the predicted and reference slope maps.  
This metric emphasizes terrain reliability: it evaluates how well slopes, ridges, and crater edges are preserved, independently of global shifts or uniform scale errors. 
In addition, we assess the structural quality of the depth maps using the Structural Similarity Index (SSIM)~\cite{ZhouWang2004}, computed between predicted and ground-truth depth maps. 
SSIM is a perceptual metric originally developed for natural images, and has been successfully adapted to depth evaluation in prior works~\cite{li2019depth,6676650}. 
It captures local geometric coherence and penalizes structural deformations, offering a complementary perspective to purely distance-based metrics. 
Finally, we propose a dedicated profile-based analysis, inspired by~\cite{Wu2016}, to assess geometric consistency along horizontal slices of the terrain. 
We extract central and evenly spaced depth profiles across the image, and compute statistics such as Mean Absolute Error (MAE), and Pearson correlation between predicted and reference profiles. 
MAE measures the absolute deviation in elevation values, while correlation reflects the similarity in terrain variations and trends. 
The combination of these two metrics provides a more complete picture: MAE captures how far the reconstruction is in absolute terms, and correlation indicates whether the relief rises and falls in a consistent pattern. 
This analysis is particularly relevant on lunar surfaces, where terrain assessment often relies on the inspection of cross-sectional elevation curves around craters or slopes.
To recover the true metric scale during inference, we apply a RANSAC-based optimal similarity transform aligning the predicted point cloud to ground truth.

\begin{figure}[h]
    \centering
    \includegraphics[width=1\linewidth]{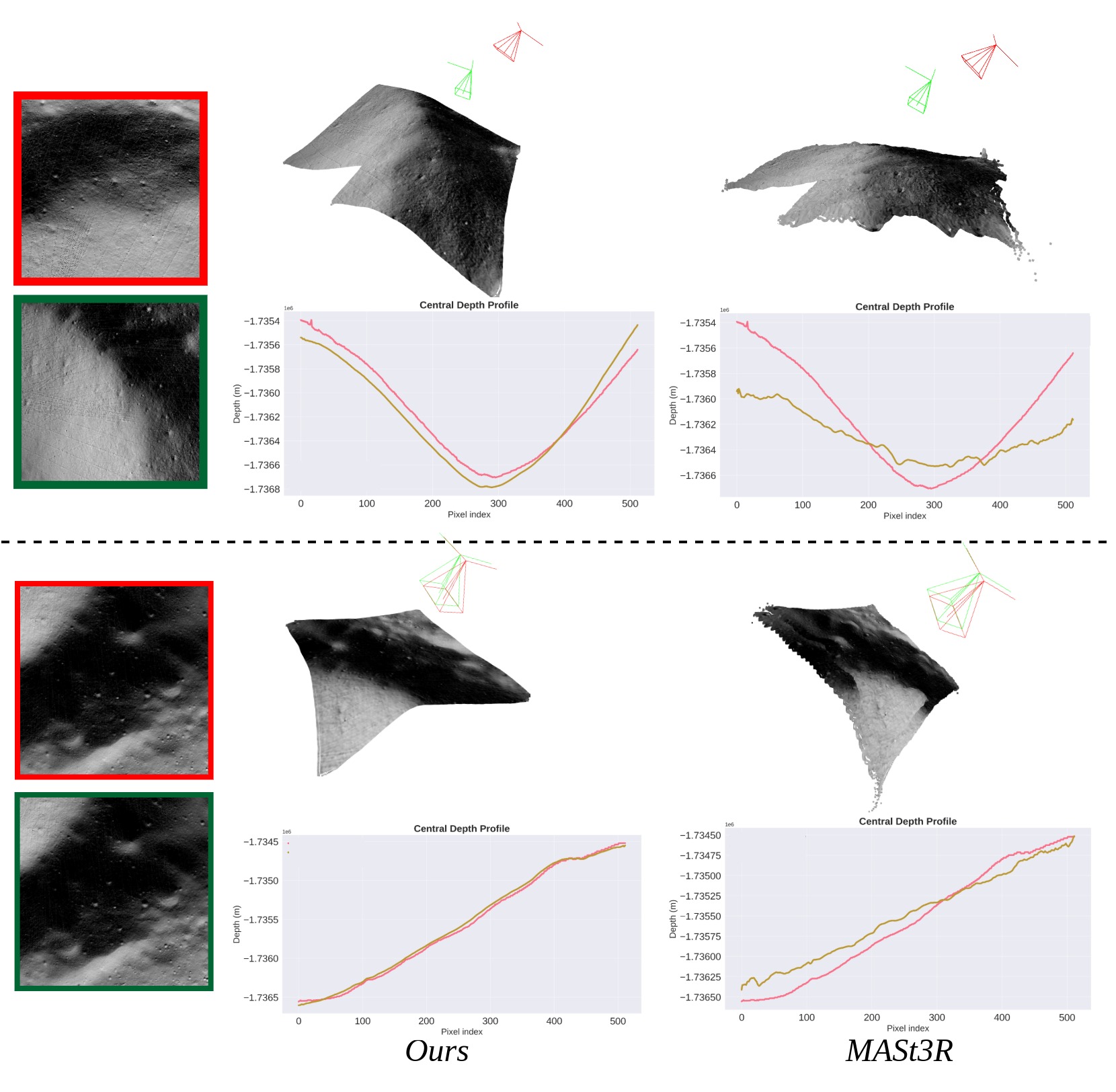}
\caption{
3D reconstruction comparison on lunar stereo pairs. 
Each row shows a stereo pair with both the reconstructed 3D point cloud and the corresponding central depth profile (ground truth in \textcolor{red}{red}, prediction in \textcolor{brown}{brown}), for \textit{MASt3R} (pretrained) and our fine-tuned model (\textit{Ours}). 
\textbf{Top}: oblique trajectory with lateral motion. 
\textbf{Bottom}: nadir view with limited parallax—a challenging case for triangulation-based SfM methods. 
\textit{Ours} yields denser and more accurate reconstructions, with better depth consistency, especially under oblique conditions. {Additional results are provided in the \textbf{supplementary material.}}}

    \label{fig:expls}
\end{figure}

\myparagraph{Results analysis} As shown in \cref{tab:metrics_split}, our method consistently outperforms MASt3R across all trajectory types and evaluation metrics. 
On absolute metrics (accuracy, completeness, and overall Chamfer error), we observe significantly lower reconstruction errors, particularly under challenging viewpoints such as oblique and random. 
This suggests that our method generalizes better to variable baselines and camera geometries. 
On structural metrics, our approach leads to substantial improvements. 
The slope correlation rises from $0.21$ to $0.80$ on Nadir sequences, and from $0.09$ to $0.76$ in the random case, indicating much better preservation of terrain morphology. 
The slope MAE also drops sharply, confirming that both the structure and steepness of the terrain are more faithfully recovered. 
SSIM and profile correlation also improve markedly (see \cref{fig:expls}), suggesting better pixel-wise depth consistency and more accurate relief along horizontal cross-sections. 
These results confirm that our approach not only achieves lower pointwise errors but also captures the underlying shape and surface of lunar terrain more robustly.

\subsection{Qualitative study with real landing images}
\label{sec:qualitative}

To further validate our approach, we extend the evaluation beyond synthetic benchmarks to real lunar imagery. 
Rather than using previously mentioned datasets, we focus on descent sequences from the Chang'E-3 mission, which offer real sensor data with richer textures and uncontrolled illumination. 
From the NavCAM video, we extract stereo pairs by cropping $512 \times 512$ patches at regular intervals.

Since no ground truth is available, we assess the results qualitatively through three criteria: (1) visual alignment of reconstructed pointmaps, (2) hillshading of the 3D output to evaluate consistency with image structure, and (3) slope map analysis. 
We observe that our fine-tuned model successfully handles nadir-like configurations (\cref{fig:qualitative-real}, left), recovering crater structure both in the hillshaded pointmap and slope map. 
In contrast, the original MASt3R model succeeds in aligning the views with correct poses, but the resulting 3D reconstruction is noisy and lacks distinctive terrain features such as craters.

In more oblique configurations (\cref{fig:qualitative-real}, right), both models estimate plausible relative poses, but only our fine-tuned version produces coherent geometry: craters and slopes remain discernible. 
MASt3R, while achieving pose alignment, yields noisy outputs with no clear relief, as confirmed by degraded hillshading and slope correlation.
The quanlitative results illustrate the generalization capacity of our proposed reconstruction network, fine-tuned on our dataset images, to real images. 
\section{Conclusion}
\label{sec:conclusion}

While recent learning-based 3D reconstruction methods have achieved impressive results, their robustness in low-texture and repetitive environments — such as the lunar surface — remains largely unexplored. 
Our work addresses this gap by introducing a physically realistic, large-scale stereo dataset specifically designed for the Moon. 
Combined with targeted fine-tuning, this enables the MASt3R model to generalize not only to our simulation data, but also to real descent imagery, demonstrating its adaptability to challenging, texture-sparse settings.

We identify two main contributions: (1) a physically grounded dataset with dense supervision, which can be extended to include real textures or alternate terrains; and (2) a demonstration that deep models like MASt3R can be successfully adapted to domains beyond their original scope—opening the door to applications on asteroids, planetary analogs, or Earth environments with poor texture.

Future work includes enriching the dataset with orthorectified real imagery, exploring fine-tuning of other architectures, improving robustness through mixed real/simulated data, and distilling the network for lightweight deployment in onboard systems. 
Finally, broader test scenarios and generalization to other planetary surfaces will further validate the framework's potential.
{
\myparagraph{Acknowledgments}
This work was carried out with the support of the European Space Agency (ESA) under contract n° 4000140461/23/NL/GLC/my.
}

{
    \small
    \bibliographystyle{ieeenat_fullname}
    \bibliography{main}
}

\end{document}


\twocolumn[{
    \maketitle

    \vspace{1.5em}

    \vspace{1em} 
}]

\begin{figure*}[!h]
    \centering
    \includegraphics[width=0.9\linewidth]{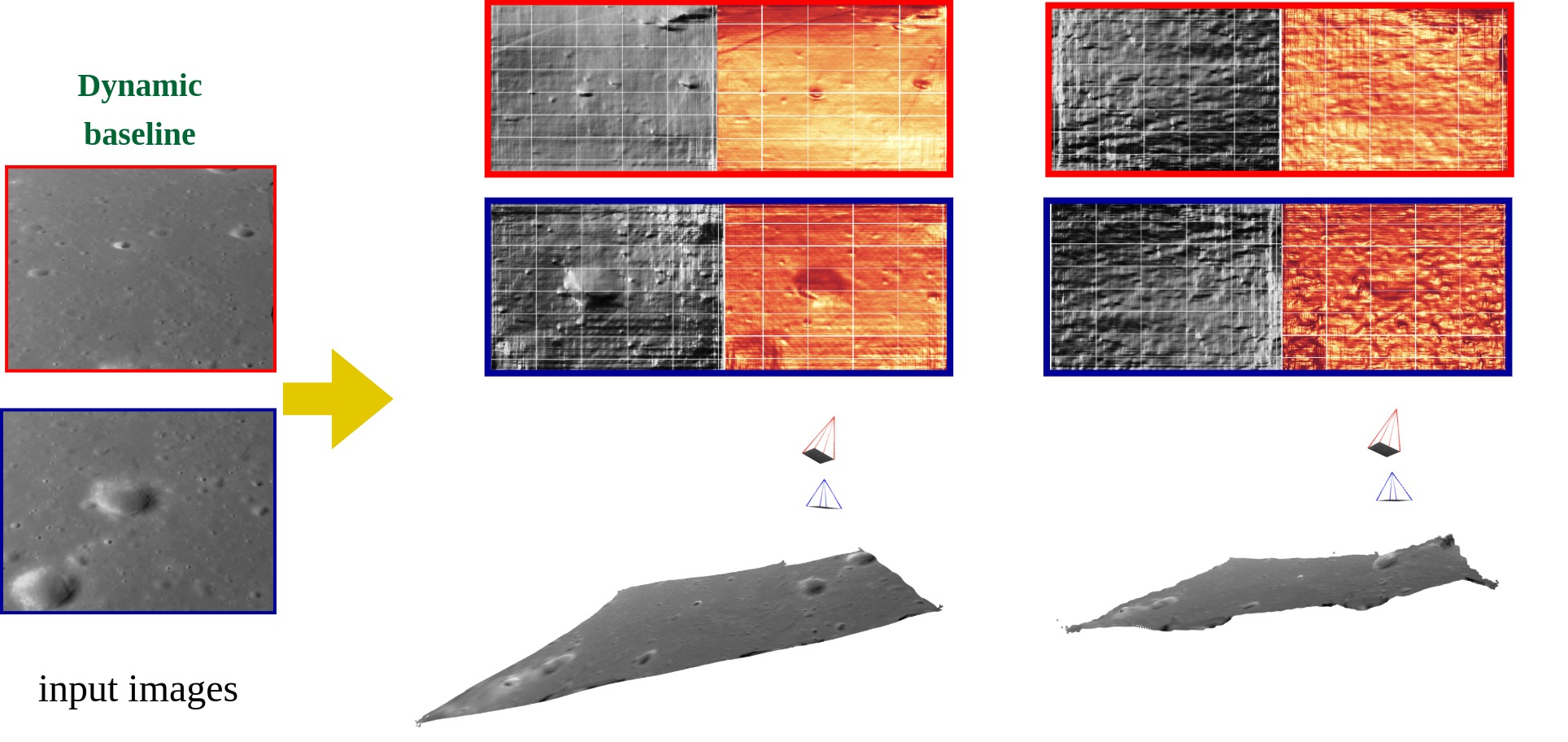}
    \caption{
Dynamic trajectory with varying altitudes and oblique viewpoints, similar to our dataset. 
For each of the two input views, from real Change'E landing (outlined in red and blue, respectively), we show the predicted hillshaded depth maps and slope maps (heat colormap), followed by the reconstructed 3D scene (bottom). 
\textbf{Left:} \textit{Ours}. \textbf{Right:} \textit{MASt3R}. 
Both models recover plausible poses, but only our method reconstructs fine-scale terrain details: crater rims and shadowed slopes appear sharper and more structurally consistent.
}
    \label{fig:real2}
\end{figure*}

\section{Additional Results on Real Lunar Landing Images}

\begin{figure*}[t]
    \centering
    \includegraphics[width=0.9\linewidth]{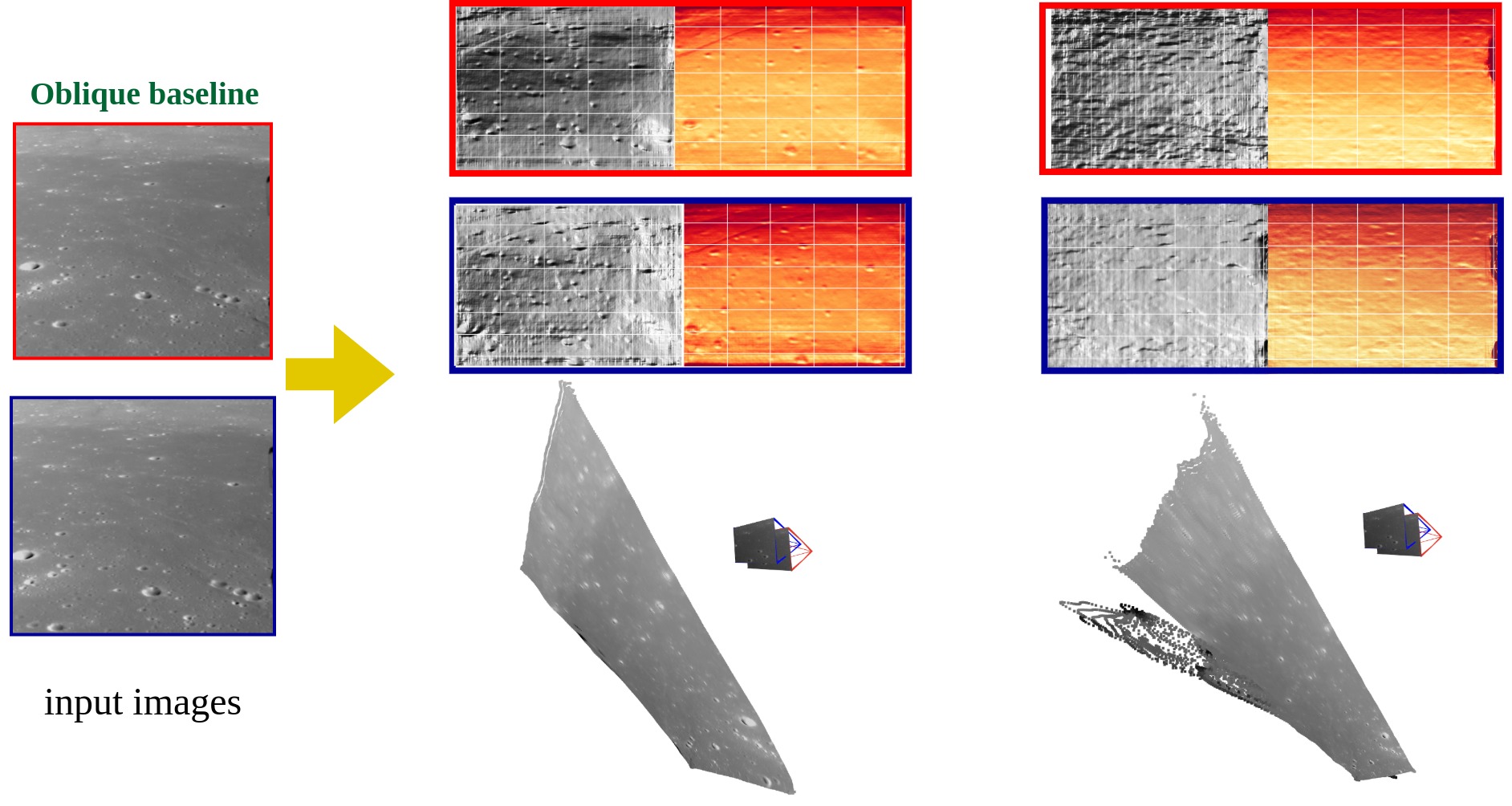}
\caption{
Oblique trajectory near touchdown. 
As in the previous example, we show for each of the two input views,  from real Change'E landing (outlined in red and blue, respectively), we show the predicted hillshaded depth maps and slope maps (heat colormap), followed by the reconstructed 3D scene (bottom).  
\textbf{Left:} \textit{Ours}. \textbf{Right:} \textit{MASt3R}. 
While both methods estimate reasonable poses, our model provides sharper gradient transitions and more coherent topographic discontinuities, particularly along crater rims and slope breaks.
}

    \label{fig:real3}
\end{figure*}

\begin{figure*}[!h]
    \centering
    \includegraphics[width=0.9\linewidth]{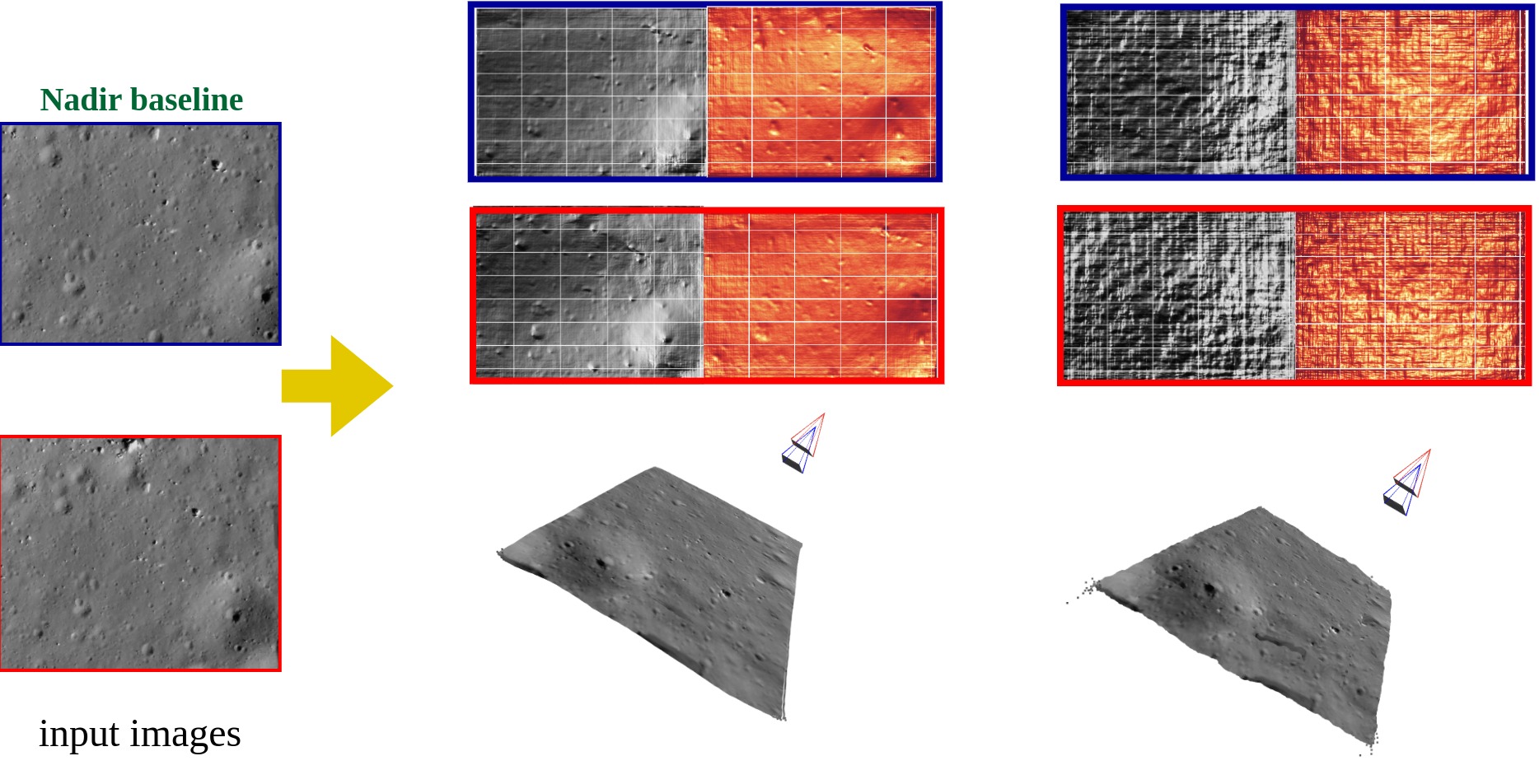}
\caption{
Vertical descent with nadir orientation. This configuration is uncommon in our dataset, where nadir views usually correspond to horizontal motion. 
For each of the two input views, from real Change'E landing  (outlined in red and blue, respectively), we display the predicted hillshaded depth maps and slope maps, followed by the full 3D reconstruction. 
\textbf{Left:} \textit{Ours}. \textbf{Right:} \textit{MASt3R}. 
Both methods estimate consistent poses, but MASt3R produces noisy geometry with limited structural detail, while our method captures more distinct terrain relief, particularly around crater rims.
}

    \label{fig:real1}
\end{figure*}

We present three additional stereo pairs from the Chang'e 4 landing to test our method on real lunar imagery: 
a dynamic case with altitude and viewpoint variation (\cref{fig:real2}), 
an oblique-oblique pair near touchdown (\cref{fig:real3}), 
and a vertical nadir descent with altitude difference (\cref{fig:real1}). 
These examples demonstrate our model’s robustness to diverse real-world configurations. 
Each figure compares the output of MASt3R (right column) with our fine-tuned model (left column), including slope maps, hillshaded depths, and 3D reconstructions.

\sisetup{group-separator = {,}}
\section{Additional Simulated Results on Challenging Cases}

\begin{figure*}[t]
    \centering
    \includegraphics[width=\textwidth]{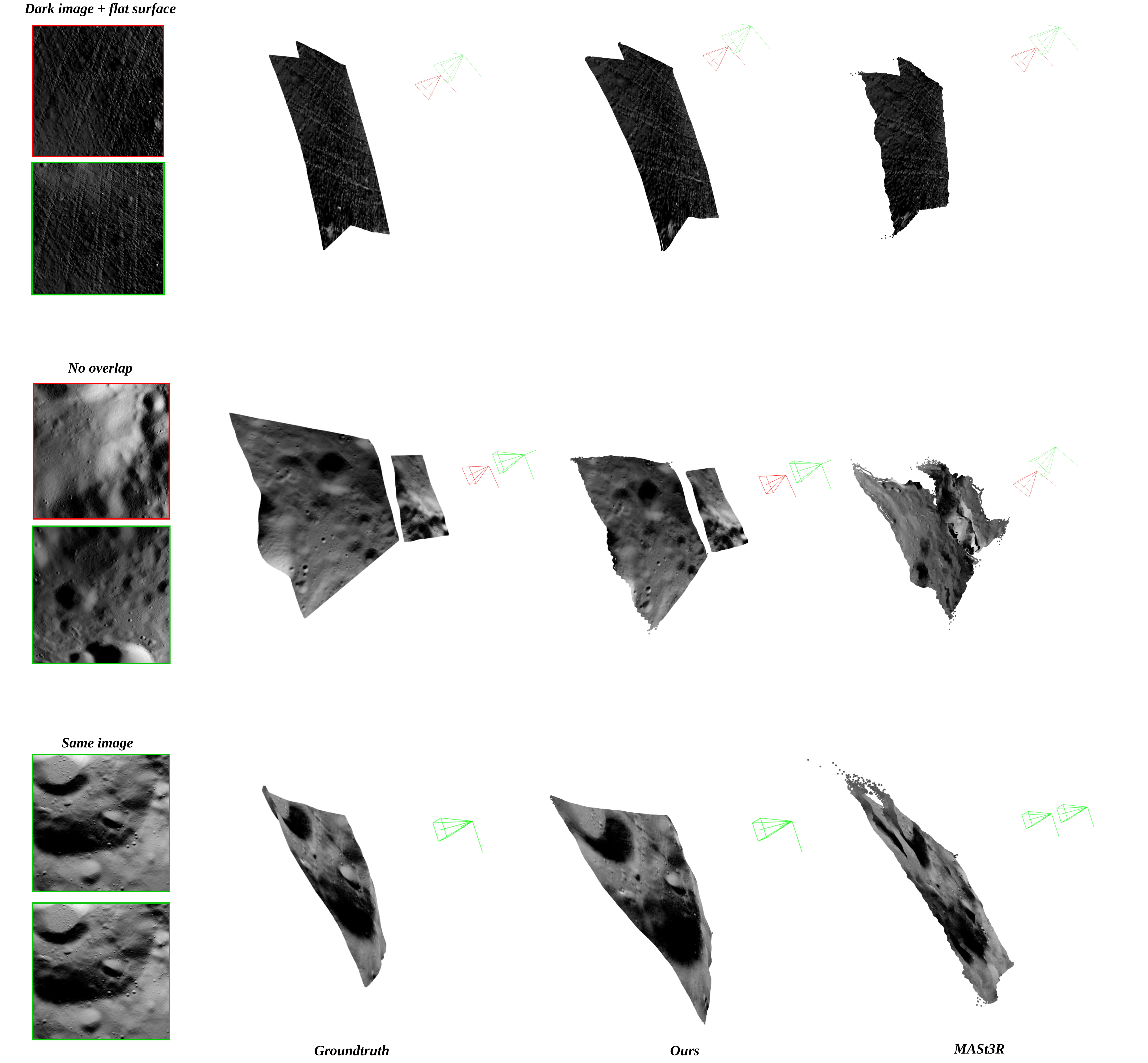}
    \caption{
Examples of 3D reconstruction in three challenging simulation cases. 
Each row shows a stereo image pair followed by reconstructed 3D point clouds from ground truth, MASt3R (middle), and our method (right). 
\textbf{Top row:} low-light images over flat terrain — both models estimate plausible poses, but only ours recovers usable geometry. 
\textbf{Middle row:} two non-overlapping images — MASt3R aligns the views incorrectly and produces collapsed geometry, while our model handles the mismatch robustly. 
\textbf{Bottom row:} a test with a pair including the same image, while MASt3R wrongly estimates a larger displacement between the views and a poor 3D reconstruction, our method correctly estimates the same pose for both views and a more reliable 3D reconstruction.
}
    \label{fig:3d_reconstruction_comparison}
\end{figure*}

We present three synthetic stereo pairs designed to stress-test both our proposed pose estimation and 3D reconstruction under adverse conditions. 
These edge cases include: (1) a low-light scene over flat terrain, (2) a pair of non-overlapping views, and (3) two nearly identical images with minimal disparity.
Despite these challenges, our model recovers accurate poses and coherent 3D structures in all three cases, whereas MASt3R struggles, producing noisy or collapsed reconstructions. 
These results are illustrated in \cref{fig:3d_reconstruction_comparison}.

